\documentclass[10pt,twocolumn,letterpaper]{article}

\usepackage{iftex}
\ifTUTeX
  \usepackage{fontspec}
  \defaultfontfeatures{Ligatures={TeX,NoCommon}}
\else
  \usepackage[T1]{fontenc}
  \usepackage[utf8]{inputenc}
  \usepackage{lmodern}
\fi
\usepackage[margin=0.75in,columnsep=0.25in]{geometry}
\usepackage{amsmath,amssymb,bm}
\usepackage{graphicx,booktabs,tabularx,multirow,array}
\usepackage{xcolor}
\usepackage[numbers,sort&compress]{natbib}
\usepackage{caption,placeins,fancyhdr,enumitem,stfloats}
\usepackage[hyphens]{url}
\usepackage[colorlinks=true]{hyperref}

\graphicspath{{../figures/}{figures/}}

\definecolor{ClayNavy}{HTML}{163E64}
\definecolor{ClayTeal}{HTML}{0F766E}
\definecolor{ClaySlate}{HTML}{334155}
\definecolor{ClayAmber}{HTML}{B45309}
\definecolor{RuleGrey}{HTML}{CBD5E1}
\definecolor{SoftBg}{HTML}{F8FAFC}

\hypersetup{
  pdftitle={Clay v1.5: An Open Sensor-Agnostic Earth Observation Foundation Model},
  pdfauthor={Bruno Sanchez-Andrade Nuno, Soumya Ranjan Mohanty, Daniel Wiesmann, Mason Grimshaw, Dan Hammer, Samuel Barrett, Konstantin Klemmer, Avery Cohn, and Laura Chen},
  pdfsubject={Geospatial Foundation Models, Sensor-Agnostic Earth Observation Embeddings, Multi-Sensor Vision Transformer, Remote Sensing Machine Learning},
  pdfkeywords={Clay v1.5, made-with-clay/Clay, sensor-agnostic Earth observation foundation model, geospatial foundation model, remote sensing embeddings, multi-sensor Vision Transformer, Masked Autoencoder, DynamicEmbedding, DOFA, Scale-MAE, AlphaEarth Foundations, TESSERA, OlmoEarth, Prithvi-EO-2.0, Copernicus-FM, Panopticon, DINOv2, Segment Anything Model, Matryoshka Representation Learning, Sentinel-2, Sentinel-1 SAR, Landsat 8/9, NAIP, LINZ, MODIS, Sentinel-3 OLCI, GEO-Bench, Copernicus-Bench, Source Cooperative, Major TOM, STAC, Apache-2.0},
  linkcolor=ClayNavy,
  citecolor=ClayTeal,
  urlcolor=ClayNavy
}

\renewcommand{\headrulewidth}{0.4pt}
\renewcommand{\headrule}{\hbox to\headwidth{\color{RuleGrey}\leaders\hrule height \headrulewidth\hfill}}
\AtBeginDocument{%
  \raggedbottom
  \setlength{\abovedisplayskip}{2.5pt}%
  \setlength{\belowdisplayskip}{2.5pt}%
  \setlength{\abovedisplayshortskip}{1.5pt}%
  \setlength{\belowdisplayshortskip}{1.5pt}%
}
\makeatletter
\renewcommand{\@seccntformat}[1]{\csname the#1\endcsname.\quad}
\renewcommand\section{\@startsection {section}{1}{\z@}%
  {-2.8pt plus -1pt minus -0.5pt}%
  {1.2pt}%
  {\normalfont\normalsize\bfseries\raggedright\color{ClayNavy}}}
\renewcommand\subsection{\@startsection{subsection}{2}{\z@}%
  {-2.2pt plus -0.8pt minus -0.5pt}%
  {0.8pt}%
  {\normalfont\normalsize\bfseries\raggedright}}
\renewcommand\paragraph{\@startsection{paragraph}{4}{\z@}%
  {1.2pt plus 0.4pt}%
  {-0.5em}%
  {\normalfont\normalsize\bfseries}}
\makeatother

\setlist[itemize]{leftmargin=*,labelsep=0.35em,itemsep=0.5pt,topsep=1pt,parsep=0pt}
\setlist[enumerate]{leftmargin=*,labelsep=0.35em,itemsep=0.5pt,topsep=1pt,parsep=0pt}

\newcolumntype{L}[1]{>{\raggedright\arraybackslash}p{#1}}
\newcolumntype{R}[1]{>{\raggedleft\arraybackslash}p{#1}}
\newcolumntype{C}[1]{>{\centering\arraybackslash}p{#1}}

\newcommand\blfootnote[1]{%
  \begingroup
  \renewcommand\thefootnote{}\footnote{#1}%
  \addtocounter{footnote}{-1}%
  \endgroup
}

\begin{document}

\twocolumn[{%
\begin{@twocolumnfalse}
\thispagestyle{plain}
{\noindent\color{ClayNavy}\rule{\textwidth}{1.5pt}\par\smallskip}
{\LARGE\bfseries\color{ClayNavy}\noindent Clay v1.5: An Open Sensor-Agnostic Earth Observation \mbox{Foundation} Model\par\smallskip}
{\normalsize\bfseries\raggedright\noindent
Bruno S\'anchez-Andrade Nu\~no$^{1,3,\ast}$,
Soumya Ranjan Mohanty$^{2}$,
Daniel Wiesmann$^{2}$,
Mason Grimshaw$^{4}$,
Dan Hammer$^{1,3,4}$,
Samuel Barrett$^{1}$,
Konstantin Klemmer$^{1,3}$,
Avery Cohn$^{1,3,4}$,
and Laura Chen$^{3,4}$\par\smallskip}
{\footnotesize\color{ClaySlate}\noindent
$^{1}$Legend AI \quad
$^{2}$Development Seed \quad
$^{3}$Clay \quad
$^{4}$Ode\par\smallskip}
{\noindent\color{RuleGrey}\rule{\textwidth}{0.6pt}\par\smallskip}
{\small\leftskip=1.8em\rightskip=1.8em\parindent=0pt
\textbf{\textcolor{ClayNavy}{Abstract.}}
Earth observation imagery serves environmental, climate, and humanitarian work at global scale, but turning it into actionable insights often takes specialist pipelines and large amounts of compute. Foundation models aim to provide one reusable backbone generalizing across tasks. Clay v1.5 is an open (Apache-2.0), $311\text{M}$-parameter ViT-L/8 masked autoencoder for Earth observation. It is conditioned on band wavelengths and ground sample distance (GSD), so one frozen checkpoint embeds optical and SAR band combinations across resolutions. It was pretrained on $72.1\text{M}$ $256\times 256$ pixel chips from seven public collections (LINZ, NAIP, Sentinel-1/2, Landsat 8/9 L1 and L2, and MODIS). The Clay project released it in November 2024 without a technical report until this one. We document the released model and its two-stage training. We evaluate the model on 43 benchmarks ($0.1\text{--}300\,\text{m}$ GSD, $2\text{--}21$ bands). We take classification and segmentation scores from an independent leaderboard and its harness, and run the tasks the leaderboard does not cover ourselves. General RGB backbones lead on RGB scenes. Multi-temporal Sentinel models lead on seasonal crop, land-cover, and tree-genus tasks, where Clay is weakest. Clay's median score on the leaderboard's 15 classification datasets is $98\%$ of the best model, and its median over all 43 benchmarks is $94\%$. With a frozen linear probe, it leads two $0.1\text{--}1\,\text{m}$ segmentation tasks; on one of them, models with multi-scale decoders score higher. We close with four lessons from two years of use, on decoders, readout and cross-sensor alignment, dimensionality reduction, and probe choice.\par\smallskip
\footnotesize\textcolor{ClaySlate}{\textbf{Keywords:} Earth observation foundation model $\cdot$ geospatial embeddings $\cdot$ multi-sensor Vision Transformer $\cdot$ masked autoencoder $\cdot$ wavelength hypernetwork $\cdot$ Apache-2.0}\\
\textbf{Artifacts:} Code: \url{https://github.com/Clay-foundation/model}~\cite{clay2024model} $\cdot$ Weights: \url{https://huggingface.co/made-with-clay/Clay}~\cite{clay2024hf} $\cdot$ Catalogs: \url{https://source.coop/clay} $\cdot$ Reproduction: \url{https://github.com/Clay-foundation/clay-v1.5-paper}\par}
{\smallskip\noindent\color{RuleGrey}\rule{\textwidth}{0.6pt}\par\medskip}
\end{@twocolumnfalse}
}]
\raggedbottom
\section{Introduction}
\label{sec:intro}

Satellite and aerial archives hold information for environmental, climate, and humanitarian applications, but their size and technical complexity often prevent their effective use. Earth observation foundation models aim to lower that barrier with general-purpose backbones that downstream tasks reuse.

The Clay project\blfootnote{$^{\ast}$Corresponding author: \href{mailto:bruno@madewithclay.org}{\texttt{bruno@madewithclay.org}}. Author contributions---Making Clay~v1.5 corpus, architecture, and training (\S\ref{sec:model}): S.R.M., D.W., M.G., and B.S.-A.N.; Evaluations, benchmarks, and applications (\S\ref{sec:comparative}--\S\ref{sec:lessons}): B.S.-A.N., D.H., S.B., K.K., M.G., L.C., and A.C.; Clay stewardship and governance: A.C., L.C., D.H., and B.S.-A.N. Apache-2.0 code and weights, CC-BY-4.0 embedding catalogs: \url{https://github.com/Clay-foundation/model}, \url{https://huggingface.co/made-with-clay/Clay}, \url{https://source.coop/clay}.} released three open Earth observation (EO) foundation model checkpoints in 2024: v0 (January 2024), v1.0 (May 2024), and v1.5 (November 2024)~\cite{clay2024model,clay2024hf}. In May 2025 the project relicensed v1.5 from OpenRAIL-M to Apache-2.0. The project never published a technical report~\cite{clay2025issue369}. This is that report: it describes the released v1.5 model and evaluates it (\S\ref{sec:model}--\S\ref{sec:lessons}). The checkpoints are distributed with public Source Cooperative GeoParquet embedding catalogs~\cite{clay2026sourcecoop}, a TorchGeo loader for these catalogs~\cite{torchgeo2026clayembeddings}, Clay backbones in TerraTorch~\cite{terratorch2026claybackbones,gomes2025terratorch}, and an ArcGIS model~\cite{esri2025claydlpk}. Major TOM also hosts v1.5 embeddings of its Core-S2L2A 249k subset~\cite{majortom2026clayv15}. Others have evaluated Clay~v1 ($768\text{-D}$ ViT-B/8) in GEO-Bench-2~\cite{simumba2026geobench2}, where it ranks first on the Core and pixel-wise subsets, and in TerraTorch, Wiratama et al.~\cite{wiratama2025indonesia}, and AlphaEarth~\cite{brown2025alphaearth}. Clay~v1.5 ($1{,}024\text{-D}$ ViT-L/8) is evaluated in OlmoEarth~\cite{herzog2026olmoearth,olmoearth2026v13}, Gong et al.~\cite{gong2026urban}, Kaushik et al.~\cite{kaushik2026flood}, and VHR-Bench~\cite{sharma2026vhrbench}. TESSERA~v2~\cite{feng2026tesserav2} also compares a Clay checkpoint but does not state which version.

Work on pretrained vision encoders~\cite{he2022mae,oquab2024dinov2} establishes that frozen backbones paired with linear probes can serve downstream tasks without fine-tuning the backbone. EO sensors span optical, thermal, hyperspectral, and synthetic aperture radar (SAR) spectra~\cite{hong2024spectralgpt,braham2025spectralearth}, cover four orders of magnitude in ground sample distance (GSD, $0.1\text{--}500\,\text{m}$)~\cite{reed2023scalemae,irvin2023usat}, and observe Earth dynamics from flood extent to seasonal crop phenology~\cite{manas2021seco,cong2022satmae}. Current EO models~\cite{zhu2026foundations,stewart2026earthembeddings} combine masked image modeling~\cite{jakubik2023prithvi,xiong2024dofa}, contrastive or self-distillation objectives~\cite{fuller2023croma,waldmann2025panopticon}, temporal sequence objectives~\cite{tseng2023presto,astruc2025anysat}, and multimodal or coordinate-aligned training~\cite{jakubik2025terramind,klemmer2025satclip}. Apart from the training objective, most EO encoders fall into two families:
\begin{enumerate}
  \item \textbf{Sentinel-anchored and fixed-schema systems.} Temporal pixel-series models such as AlphaEarth Foundations and TESSERA~\cite{feng2026tessera} summarize multi-date observations into $10\,\text{m}$ pixel vectors ($64\text{--}128\text{-D}$) distributed as annual composites. Fixed-schema encoders (AnySat, SkySense~v2~\cite{zhang2025skysensev2}, Prithvi-EO-2.0~\cite{szwarcman2026prithvi2}, TerraMind, Galileo~\cite{tseng2025galileo}, and OlmoEarth) attach dedicated stems to predefined sensor schemas. These models do best on $10\text{--}30\,\text{m}$ phenology-driven tasks. They either ship precomputed rasters without model weights (AlphaEarth) or bind inference to fixed band sets.
  \item \textbf{Physically conditioned, sensor-agnostic encoders.} Models such as Clay, Panopticon, Copernicus-FM~\cite{wang2025copernicusfm}, and FlexiMo~\cite{li2026fleximo} generate patch projections, positional encodings, or both from wavelength and GSD at run time, following Scale-MAE, USat, ChannelViT~\cite{bao2024channelvit}, and DOFA. One frozen checkpoint then embeds unseen instruments and sub-meter to kilometer imagery without retraining.
\end{enumerate}

\section{The Model}
\label{sec:model}

\begin{figure*}[!t]
\centering
\includegraphics[width=0.95\textwidth]{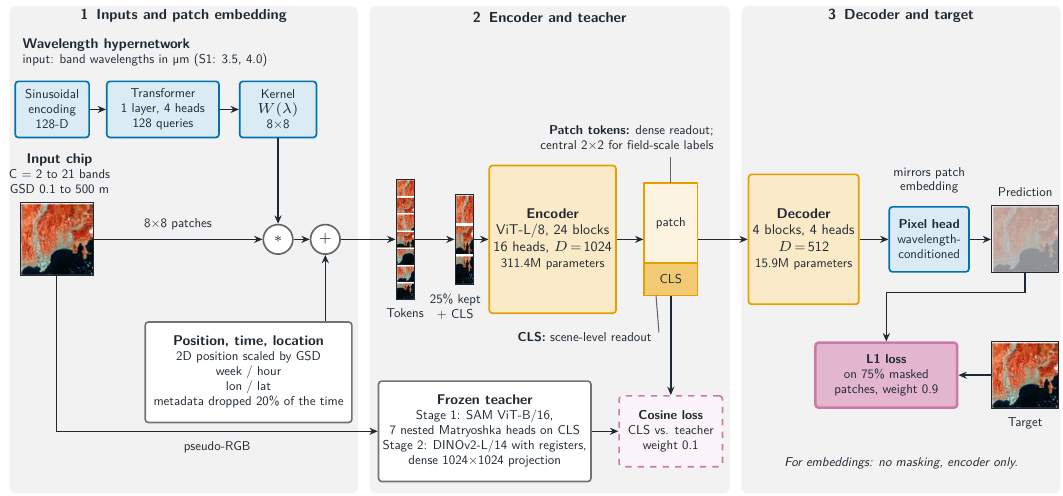}
\caption{\textbf{Clay~v1.5 architecture.} (1)~Band wavelengths pass through a hypernetwork (sinusoidal encoding, one transformer layer with 4 heads and 128 queries) that generates the $8\times 8$ patch-embedding kernel, so chips with any band set can be embedded; position, time, and location are added. (2)~A ViT-L/8 encoder ($311.4\text{M}$ parameters) processes the $25\%$ of tokens left unmasked plus a \texttt{CLS} token. \texttt{CLS} is distilled from a frozen teacher (SAM ViT-B/16 in Stage~1, DINOv2-L/14 in Stage~2). (3)~A decoder ($15.9\text{M}$ parameters) reconstructs the masked patches. Loss weights are $0.9$ ($L_1$) and $0.1$ (cosine). Embeddings are computed with the encoder alone.}
\label{fig:arch}
\end{figure*}

Clay~v1.5 was pretrained on $72.1\text{M}$ $256\times 256$ chips ($33.8\,\text{TB}$) across seven open STAC collections spanning $0.5\text{--}500\,\text{m}$ GSD (Table~\ref{tab:corpus}, Figure~\ref{fig:arch}).

\subsection{Pretraining Data}
\label{sec:corpus}

All imagery came from openly licensed public archives. We ingested Sentinel-1 RTC and MODIS MOD09A1 from \textbf{Microsoft Planetary Computer}, which also served the NAIP STAC index. We copied Sentinel-2 L2A (Element~84 Earth Search), NAIP, Landsat~8/9 Collection~2 L1/L2, LINZ, and the $80\,\text{m}$ ESA WorldCover stratification layer from \textbf{AWS Open Data} S3 buckets.

We stratified global scenes from $2{,}728$ unique MGRS tiles ($100\times 100\,\text{km}$, 2018--2023) by $80\,\text{m}$ ESA WorldCover~2021~\cite{zanaga2022worldcover} to oversample human-modified and ecologically diverse landscapes. Built-up, agricultural, and class-diverse strata together account for $77.5\%$ of tile draws; built-up land is drawn at ${\sim}40\times$ its $<1\%$ share of global land. Natural land-cover classes share the remainder. NAIP samples populated places, protected areas, infrastructure, hydrography, and $4{,}000$ random CONUS sites. LINZ samples $50\%$ of New Zealand regional catalogs, and MODIS samples $233$ sinusoidal tiles monthly (2018--2023).

\texttt{stacchip}~\cite{clay2024stacchip} was built to cut $256\times 256$ windows on the fly from Cloud-Optimized GeoTIFFs (COGs) indexed in GeoParquet. But random HTTP/S3 range reads across 2--10 band files per chip left the GPUs waiting on I/O and GDAL decompression. We therefore mirrored selected scenes to S3 and used AWS Batch to pre-pack chips into compressed $128$-chip \texttt{.npz} datacubes per sensor. We split each file into four $32$-chip sub-cubes staged on Amazon FSx for Lustre (\path{/fsx}), so each \texttt{DataLoader} worker reads a contiguous $32$-chip block in a single I/O operation. The train/val split is $80/20$.

\begin{table}[!b]
\centering
\caption{Clay~v1.5 model card and pretraining corpus ($256\times 256$ chips, $33.8\,\text{TB}$; native NAIP $0.6\text{--}1.0\,\text{m}$). Nominal GSD $g$ is passed to the encoder; each $8\times 8$ token covers $8g\times 8g$ ($256g$ chip width). Chip counts are the inventory in the upstream release notes. Training draws equally from the seven collections ($14.3\%$ of steps each), so chips in small collections are drawn more often: a MODIS chip is drawn $7.6$ times as often as it would be if all chips were sampled uniformly ($\times 7.6$), and a NAIP chip half as often ($\times 0.5$).}
\label{tab:corpus}
\scriptsize
\setlength{\tabcolsep}{2.4pt}
\renewcommand{\arraystretch}{0.92}
\begin{tabularx}{\columnwidth}{>{\raggedright\arraybackslash}Xrrrr}
\toprule
\multicolumn{5}{p{0.96\columnwidth}}{\textbf{Clay v1.5} (\texttt{made-with-clay/Clay}~\cite{clay2024hf}) $\cdot$ ViT-L/8 ($311.4\text{M}$ params, $327.3\text{M}$ w/ decoder; vs.\ $95.3\text{M}$ v0, $92.1\text{M}$ v1.0) $\cdot$ Apache-2.0~\cite{clay2024model}} \\
\midrule
Collection (Source) & Bands & GSD & Token & \shortstack[r]{Chips (M)\\ ($\times$ chip share)} \\
\midrule
LINZ Aerial (AWS) & 3 RGB & 0.5\,m & 4\,m & 3.30 ($\times 3.1$) \\
USDA NAIP (PC/AWS) & 4 RGBN & 1.0\,m & 8\,m & 20.98 ($\times 0.5$) \\
Sentinel-2 L2A (AWS) & 10 VSWIR & 10\,m & 80\,m & 18.68 ($\times 0.6$) \\
Sentinel-1 RTC (PC) & 2 SAR & 10\,m & 80\,m & 16.13 ($\times 0.6$) \\
Landsat 8/9 L1 (AWS) & 6 VSWIR & 30\,m & 240\,m & 5.83 ($\times 1.8$) \\
Landsat 8/9 L2 (AWS) & 6 VSWIR & 30\,m & 240\,m & 5.79 ($\times 1.8$) \\
MODIS MOD09A1 (PC) & 7 VSWIR & 500\,m & 4\,km & 1.35 ($\times 7.6$) \\
\midrule
\textbf{Total / Span} & \textbf{2--10} & \textbf{0.5--500\,m} & \textbf{4\,m--4\,km} & \textbf{72.07} \\
\bottomrule
\end{tabularx}
\end{table}

\begin{table*}[p]
\centering
\caption{Clay~v1.5 against EO and general-purpose models on $43$ benchmarks. \texttt{Best other EO} and \texttt{Best general} give the best EO foundation model other than Clay and the best general-purpose model in each row, with the model tag in brackets. \textbf{\% Best} is Clay divided by the best value in the row, and bold marks that value. A slash separates probe / fine-tuning (Panel B, where the row says so) or \texttt{10-shot / Max} (AlphaEarth/TESSERA rows); each side is bolded on its own and \% Best takes the better ratio. $\pm$ in Panel~A is the leaderboard's reported uncertainty, or the standard deviation over seeds in the segmentation rows.}
\label{tab:master_bench}
\fontsize{6.5}{7.4}\selectfont
\setlength{\tabcolsep}{1.2pt}
\setlength{\aboverulesep}{0pt}
\setlength{\belowrulesep}{0pt}
\renewcommand{\arraystretch}{1.08}
\newcommand{\slsh}{{/}}
\begin{tabular*}{0.99\textwidth}{@{\extracolsep{\fill}}ll r r l l@{}}
\toprule
Benchmark & Sensor and metric & \textbf{Clay v1.5} & \textbf{\% Best} & Best other EO & Best general \\
\midrule
\multicolumn{6}{@{}l@{}}{\emph{Panel A: SatML leaderboard. Classification, linear probe, 15 core datasets}} \\
\texttt{aid} & Aerial RGB, Acc & 92.8 $\pm$ 1.3 & 94\% & 98.0$^\dagger$ (FarSLIP) & \textbf{98.2} (DINOv3-L, EG2) \\
\texttt{benv2} & Sentinel-1 + 2, micro-mAP & 83.7 $\pm$ 0.6 & 97\% & \textbf{86.4} (Olmo-v1.2-B) & 81.6 (DINOv3-L) \\
\texttt{eurosat} & Sentinel-2, Acc & 98.7 $\pm$ 0.3 & 100\% & \textbf{99.1} (Olmo-v1-B) & 97.9 (ViT-L/16) \\
\texttt{eurosat-spatial} & Sentinel-2, Acc & 96.8 $\pm$ 0.5 & 99\% & \textbf{98.3} (Olmo-v1-L) & 96.1 (ConvNeXt-L) \\
\texttt{forestnet} & Sentinel-2, Acc & 58.3 $\pm$ 3.0 & 98\% & \textbf{59.3} (DEO-B-RGB) & 57.0 (ViT-L/16) \\
\texttt{m-bigearthnet} & Sentinel-2, micro-mAP & 73.1 $\pm$ 1.4 & 93\% & \textbf{78.4} (Olmo-v1.2-B) & 70.9 (DINOv3-L) \\
\texttt{m-brick-kiln} & Sentinel-2, Acc & \textbf{98.0} $\pm$ 0.9 & 100\% & 97.7 (TM-L-RGB) & 97.9 (SwinV2-B) \\
\texttt{m-eurosat} & Sentinel-2, Acc & 96.8 $\pm$ 1.1 & 98\% & \textbf{98.6} (Olmo-v1-L) & 96.4 (DINOv3-L) \\
\texttt{m-forestnet} & Landsat, Acc & 57.5 $\pm$ 3.1 & 98\% & \textbf{58.9} (DEO-B-RGB) & 56.3 (Swin-L) \\
\texttt{m-pv4ger} & Aerial RGB, Acc & 96.1 $\pm$ 1.1 & 98\% & 97.1 (DOFA-L) & \textbf{97.9} (ConvNeXt-L) \\
\texttt{m-so2sat} & Sentinel-1 + 2, Acc & 73.0 $\pm$ 2.7 & 98\% & \textbf{74.4} (Olmo-v1.1-B) & 67.2 (DINOv3-L) \\
\texttt{resisc45} & Aerial RGB, Acc & 88.1 $\pm$ 0.7 & 92\% & 95.2$^\dagger$ (FarSLIP) & \textbf{95.8} (DINOv3-L) \\
\texttt{so2sat} & Sentinel-1 + 2, Acc & 73.0 $\pm$ 2.9 & 98\% & \textbf{74.8} (Olmo-v1-L) & 67.2 (DINOv3-L) \\
\texttt{treesatai} & Aerial + S1/S2, micro-mAP & 63.0 $\pm$ 1.4 & 93\% & \textbf{68.1} (DINOv3-SAT) & 65.8 (DINOv3-L) \\
\texttt{ucmerced} & Aerial RGB, Acc & 96.7 $\pm$ 1.6 & 97\% & \textbf{99.3} (DOFA-L) & 99.0 (EfficientNetV2-L) \\
\multicolumn{6}{@{}l@{}}{\emph{Segmentation, UPerNet head}} \\
\texttt{sen1floods11-s2} & Sentinel-2, mIoU & \textbf{89.1} $\pm$ 0.1 & 100\% & 89.0 (Olmo-v1.2-B) & 78.7 (SwinV2-L) \\
\texttt{sen1floods11-s1} & Sentinel-1, mIoU & 79.6 $\pm$ 0.3 & 97\% & \textbf{82.2} (Olmo-v1.2-B) & 81.7 (SwinV2-L) \\
\texttt{burn\_scars} & HLS, mIoU & 88.2 $\pm$ 0.4 & 94\% & \textbf{94.0} (Pr.-2) & 90.0 (DINOv3-L) \\
\midrule
\multicolumn{6}{@{}l@{}}{\emph{Panel B: Benchmarks used by peers (our runs). OlmoEarth v1.3, GEO-Bench and Copernicus-Bench tasks, linear probe or fine-tune}} \\
\texttt{m-pv4ger-seg} & Aerial RGB $0.1\,\text{m}$ (mIoU) & \textbf{84.3} & 100\% & 80.0$^\natural$ (Olmo) & 77.9 (DINOv2) \\
\texttt{m-chesapeake} & NAIP RGBN $1.0\,\text{m}$ (mIoU) & \textbf{66.9} & 100\% & 59.9 (Pan.) & 60.2 (DINOv2) \\
\texttt{Biomass-S3}$^\ast$ & S3 OLCI $300\,\text{m}$, 21b ($R^2$) & 0.538 & 91\% & \textbf{0.594}$^\natural$ (Olmo) & 0.451 (DINOv2) \\
\texttt{LC100Cls-S3}$^{\ast\ddagger}$ & S3 OLCI $300\,\text{m}$, 21b (mAP) & 71.6 & 94\% & \textbf{76.4} (Pan.) & 69.1 (DINOv2) \\
\texttt{EuroSAT-S1}$^\ddagger$ & S1 land use (OA) & 85.9 & 97\% & \textbf{88.7} (Pan.) & 84.0 (DINOv2) \\
\texttt{Vessel Detect.\ (S1)} & S1 vessels (FT $F_1$) & 77.5$^\S$ & 97\% & \textbf{79.6} (TM-B) & n/a \\
\texttt{Vessel Detect.\ (L8)} & Landsat-8 vessels (FT $F_1$) & 66.4$^\S$ & 89\% & \textbf{74.9} (Pan.) & n/a \\
\texttt{Solar Farm Detect.} & S2 10b solar (FT mIoU) & 81.8 & 93\% & \textbf{87.5} (Olmo) & 70.7 (DINO) \\
\texttt{MADOS} & S2 debris (LP / FT mIoU) & 52.4 \slsh{} 68.8 & 92\% & \textbf{68.4} \slsh{} \textbf{74.6} (Gal./Olmo) & 59.7 \slsh{} 64.7 (DINO) \\
\texttt{m-cashewplant} & S2 plantations (LP / FT mIoU) & 20.6 \slsh{} 70.7$^\S$ & 87\% & 51.1 \slsh{} \textbf{81.3} (Pr.-2/TM) & \textbf{54.1} \slsh{} 80.6 (DINO) \\
\multicolumn{6}{@{}l@{}}{\emph{AlphaEarth / TESSERA suite, 15 tasks, 10-shot / Max}} \\
\texttt{Africa crop mask} & Cropland mask (B.Acc) & 0.77 \slsh{} 0.87 & 95\% & \textbf{0.82} \slsh{} \textbf{0.92}$^\diamond$ (Olmo) & 0.77 \slsh{} 0.88 (DINOv2) \\
\texttt{ASTER emissivity} & Thermal emissivity ($R^2$) & 0.75 \slsh{} 0.84 & 98\% & \textbf{0.80} \slsh{} \textbf{0.86} (Olmo) & 0.49 \slsh{} 0.64 (DINOv2) \\
\texttt{LCMAP land cover} & CONUS land cover (B.Acc) & 0.67 \slsh{} 0.74 & 91\% & \textbf{0.74} \slsh{} \textbf{0.84}$^\diamond$ (Olmo/AEF) & 0.55 \slsh{} 0.66 (DINOv2) \\
\texttt{LCMAP land use} & CONUS land use (B.Acc) & 0.66 \slsh{} 0.74 & 88\% & \textbf{0.75} \slsh{} \textbf{0.89}$^\diamond$ (Olmo) & 0.54 \slsh{} 0.57 (DINOv2) \\
\texttt{LCMAP cover change} & Cover change (B.Acc) & 0.56 \slsh{} 0.58 & 90\% & \textbf{0.62} \slsh{} \textbf{0.78}$^\diamond$ (AEF) & 0.50 \slsh{} 0.51 (DINOv2) \\
\texttt{LCMAP use change} & Use change (B.Acc) & 0.53 \slsh{} 0.56 & 95\% & \textbf{0.56} \slsh{} \textbf{0.79}$^\diamond$ (TES/AEF) & 0.49 \slsh{} 0.45 (DINOv2) \\
\texttt{Oil palm (Descals)} & Oil palm extent (B.Acc) & 0.65 \slsh{} 0.71 & 86\% & \textbf{0.76} \slsh{} \textbf{0.84} (Olmo) & 0.65 \slsh{} 0.79 (DINOv2) \\
\texttt{GLanCE land cover} & Global land cover (B.Acc) & 0.49 \slsh{} 0.59 & 86\% & \textbf{0.57} \slsh{} \textbf{0.71}$^\diamond$ (Olmo) & 0.38 \slsh{} 0.44 (DINOv2) \\
\texttt{LUCAS land use} & Europe land use (B.Acc) & 0.23 \slsh{} 0.26 & 77\% & \textbf{0.30} \slsh{} \textbf{0.39}$^\diamond$ (Olmo/AEF) & 0.18 \slsh{} 0.20 (DINOv2) \\
\texttt{LUCAS land cover} & Europe land cover (B.Acc) & 0.18 \slsh{} 0.22 & 55\% & \textbf{0.33} \slsh{} \textbf{0.41}$^\diamond$ (TES) & 0.10 \slsh{} 0.12 (DINOv2) \\
\texttt{Canada crops (crs.)} & Coarse crop types (B.Acc) & 0.35 \slsh{} 0.48 & 65\% & \textbf{0.57} \slsh{} \textbf{0.74}$^\diamond$ (Olmo) & 0.24 \slsh{} 0.29 (DINOv2) \\
\texttt{Canada crops (fine)} & Fine crop types (B.Acc) & 0.35 \slsh{} 0.45 & 68\% & \textbf{0.54} \slsh{} \textbf{0.66}$^\diamond$ (Olmo) & 0.17 \slsh{} 0.22 (DINOv2) \\
\texttt{Ethiopia crops} & Smallholder crops (B.Acc) & 0.31 \slsh{} 0.32 & 84\% & \textbf{0.37} \slsh{} \textbf{0.69}$^\diamond$ (TES/AEF) & 0.27 \slsh{} 0.33 (DINOv2) \\
\texttt{US tree genera} & Tree genera, 39c (B.Acc) & 0.07 \slsh{} 0.09 & 54\% & \textbf{0.13} \slsh{} \textbf{0.23}$^\diamond$ (Olmo) & 0.07 \slsh{} 0.07 (DINOv2) \\
\texttt{OpenET monthly ET} & S2 flux ($R^2$) & 0.12 \slsh{} 0.45 & 80\% & \textbf{0.15} \slsh{} \textbf{0.58}$^\diamond$ (Olmo/AEF) & \text{--}0.27 \slsh{} 0.09 (DINOv2) \\
\midrule
\textbf{Median, all panels} & 43 rows & 1 checkpoint & \textbf{94\%} & 11 model families & 7 model families \\
\bottomrule
\end{tabular*}
\par\smallskip
\begin{minipage}{0.99\textwidth}
\scriptsize\setlength{\parskip}{1.5pt}
\textbf{Sources.} Panel~A is read from the SatML leaderboard~\cite{taylorgeospatial2026satml} at harness commit \texttt{83ca5cb} (2026-11-16), with the best band set and normalization per model; Clay's $\pm$ there is the leaderboard's, for one run. In the segmentation rows, Clay is read from the harness's own results at commit \texttt{37c0e3e} (2026-12-01), as the mean and standard deviation of 3 seeds with a frozen encoder, while the six leaderboard cells beside them are each model's best single run, five of them with LoRA or full fine-tuning. Clay~v1 Base, not offered as a comparator, scores $89.4$ on \texttt{sen1floods11-s2}. In Panel~B, \texttt{Best other EO} is the maximum over DOFA-L, OlmoEarth~v1.3, the best other peer, TESSERA, and AlphaEarth, and \texttt{Best general} is DINOv2 or DINOv3; the peer cells of \texttt{Vessel Detect.}, \texttt{Solar Farm Detect.}, \texttt{MADOS}, and \texttt{m-cashewplant} are from~\cite{olmoearth2026v13}. The last row gives the median \% Best over all rows and the number of model families behind each comparison column, where versions and sizes of one model count once.\par
\textbf{Marks.} $^\dagger$Our own run in the leaderboard's harness, of a model the leaderboard does not list. $^\ast$Sentinel-3 OLCI, a 21-band sensor not seen in pretraining, with zero-decoder, image-only probes; \texttt{Biomass-S3} is at its native $96\,\text{px}$ (chip $R^2{=}0.790$) and \texttt{LC100Cls-S3} reports macro mAP. $^\ddagger$Chips under $224\,\text{px}$ are resized to $224\times 224$ for the linear probe. $^\natural$Cross-sensor transfer: RGB/NIR bands mapped into \texttt{SENTINEL2\_L2A} \texttt{BandSet 0} with \texttt{MaskValue.MISSING}, as in~\cite{olmoearth2026v13}. $^\S$Single seed: \texttt{Vessel Detect.} at $48{,}000$ steps and \texttt{m-cashewplant} at 100 epochs with the better of two learning rates. Validation had leveled off except on \texttt{Vessel Detect.\ (S1)}. A three-seed run of \texttt{Vessel Detect.\ (S1)} at $6{,}000$ steps gives $63.1$. \cite{olmoearth2026v13} reports $79.9$, $70.7$, and $73.9$ for Clay on these three rows. It has no DINO result for the two \texttt{Vessel Detect.} rows (n/a), and we did not run one. $^\diamond$Published $10\,\text{m}$ 12-month embeddings: \texttt{Ridge} with $\lambda{=}0$ on the 8 tasks shared with~\cite{olmoearth2026v13}, and the remaining \texttt{Max} cells from~\cite{brown2025alphaearth} Fig.~1A and~\cite{feng2026tesserav2} Table~2.\par
\textbf{Panel~B.} \texttt{DINOv2} uses 3 bands on tasks with more than 3 bands and $[\text{VV},\text{VH},(\text{VV}{+}\text{VH})/2]$ on Sentinel-1.\par
\textbf{AlphaEarth/TESSERA rows.} Single-date ViTs use the central $160\,\text{m}$ patch readout (\texttt{Clay}, \texttt{OlmoEarth}, \texttt{Pr.-2}) or a $224\,\text{px}$ pool (\texttt{DOFA-L}, \texttt{DINOv2}), unless marked $^\diamond$. Unmarked \texttt{Max} cells use the labels available on the matched rows. Unmarked \texttt{TESSERA} / \texttt{AEF} cells use $10\,\text{m}$ v1.1~\cite{feng2026tessera} / $320\,\text{m}$ annual rasters; the published \texttt{AEF} \texttt{OpenET} value ($0.58$) is aligned to the label month.\par
\textbf{Model tags.} \texttt{Olmo-v<V>-<S>}: OlmoEarth version V, size S (B base, L large); \texttt{Olmo}: v1.3. \texttt{DOFA-L}: DOFA Large~\cite{xiong2024dofa}. \texttt{Pr.-2}: Prithvi-EO-2.0 (600M in Panel~A)~\cite{szwarcman2026prithvi2}. \texttt{TM}, \texttt{TM-<S>}: TerraMind v1, size S, with \texttt{-RGB} for RGB-only input~\cite{jakubik2025terramind}; likewise \texttt{DEO-B-RGB}: DEO Swin-B~\cite{wolf2026deo}. \texttt{Gal.}: Galileo~\cite{tseng2025galileo}. \texttt{Pan.}: Panopticon. \texttt{TES}: TESSERA. \texttt{AEF}: AlphaEarth. \texttt{DINOv3-L}, \texttt{DINOv3-SAT}: DINOv3 ViT-L/16 and the satellite-pretrained ViT-L/16~\cite{simeoni2025dinov3}. \texttt{DINOv2}: DINOv2-L~\cite{oquab2024dinov2}. \texttt{DINO}: DINOv2 or DINOv3, as reported. \texttt{EG2}: EmbeddingGemma~2~\cite{schechtervera2026embeddinggemma2}, and \texttt{FarSLIP}: FarSLIP ViT-B/16~\cite{li2025farslip}; both are our runs in the same harness on RGB input. Other tags are leaderboard model names.\par
\textbf{Datasets.}~\cite{mayer2022pv4ger,robinson2019chesapeake,buchhorn2020lc100,santoro2024biomasscci,melton2022openet,wang2025fgmae,lee2021brickkiln,zhu2020so2sat,irvin2020forestnet,kerner2024africacrop,kerner2024africacropdata,hulley2015astergedd,brown2020lcmap,pengra2023lcmapref,descals2021oilpalm,descals2024oilpalm,stanimirova2023glance,dandrimont2020lucas,blasch2024ethct2020,garnot2021pastis,jin2021cashew,kikaki2024mados,sumbul2019bigearthnet,sumbul2021bigearthnetmm,bonafilia2020sen1floods11,helber2019eurosat,cheng2017resisc45,yang2010ucmerced}.
\end{minipage}
\end{table*}

\subsection{Architecture}
\label{sec:arch}

Clay~v1.5 is a ViT-Large masked autoencoder~\cite{dosovitskiy2021vit,he2022mae} with $24$ pre-norm transformer blocks ($D=1024$, $16$ heads). Its $8\times 8$ patch stride yields $L = (H/8)(W/8)$ patch tokens: $1{,}024$ tokens per $256\times 256$ chip, $4\times$ the token density of $16\times 16$ ViTs. A 4-block, 4-head decoder ($D_{\mathrm{dec}}=512$) re-adds the position and metadata encodings to the visible and mask tokens and ends in a wavelength-conditioned pixel head that mirrors the patch embedding. Three conditioning modules adapt the backbone across sensors, scales, and observation contexts:
\begin{enumerate}
  \item \textbf{Wavelength hypernetwork (\texttt{Dynamic\-Embedding}).} A standard ViT binds its input channels to a fixed sensor schema and must be retrained whenever a mission adds channels or shifts spectral windows. \texttt{Dynamic\-Embedding}, adapted from DOFA, generates the $8\times 8$ patch-projection kernel $\bm{W}(\bm{\lambda}) \in \mathbb{R}^{1024 \times C \times 8 \times 8}$ and bias $\bm{b}(\bm{\lambda}) \in \mathbb{R}^{1024}$ on the fly from band center wavelengths $\bm{\lambda} = (\lambda_1,\dots,\lambda_C)$ in micrometers ($\mu\text{m}$). It uses a $128\text{-D}$ sinusoidal encoding, a residual MLP, and a $1$-layer, $4$-head transformer over $128$ learnable weight queries. Queries are initialized with standard deviation $0.02$; generated kernels are multiplied by $0.02$. C-band Sentinel-1 $\gamma^0$ in dB uses pseudo-wavelengths $\lambda_{\mathrm{VV}}=3.5\,\mu\text{m}$ and $\lambda_{\mathrm{VH}}=4.0\,\mu\text{m}$. Because $\bm{W}(\lambda)$ varies smoothly across visible, NIR, and SWIR passbands, shifted band centers across Sentinel-2, Landsat~8/9, NAIP, or unseen $21$-band Sentinel-3 OLCI map onto shared spectral filters at run time. On $320$ held-out Sentinel-2 scenes, dropping from $10$ bands to the $6$ Landsat-equivalent bands retains $95.3\%$ Top-1 retrieval ($\cos=0.940$ on $\overline{\bm{Z}}_{\mathrm{patch}}$) and $R^2=0.807$ (vs.\ $0.839$ on $10$ bands). Every input band therefore needs a center wavelength, so Clay is not suited to derived layers such as DEMs or NDVI.
  \item \textbf{GSD-scaled 2D positional encoding.} An $8\times 8$ token spans $4\times 4\,\text{m}$ on $0.5\,\text{m}$ LINZ orthoimagery but $4\times 4\,\text{km}$ on $500\,\text{m}$ MODIS. Following Scale-MAE, channels $0\text{--}1015$ scale 2D token coordinates by nominal GSD $g$, so self-attention allocates receptive fields by physical ground distance rather than by pixel count alone.
  \item \textbf{Time and location encoding.} The same reflectance means different things across seasons, sun angles, and biomes. Channels $1016\text{--}1023$ append $[\sin,\cos]$ encodings of ISO week, UTC hour, and $(\text{lon},\text{lat})$ coordinates (\texttt{1020..1023}).
\end{enumerate}

\subsection{Training, Compute, and License}
\label{sec:lineage}

A custom sampler and collate function cycle round-robin through the seven sensors, one sensor per step on every GPU ($32$ chips per GPU), and repeat chips from smaller collections so each gets $1/7$ of the steps. We chose this over a curriculum that trains on one sensor for long stretches. Because each mini-batch comes from a single sensor, masked pixel reconstruction alone could leave each sensor in its own embedding space. To favor learning of localized Earth patterns, Clay~v1.5 always masks $75\%$ of patch tokens, a ratio we kept from Clay~v0, where reconstructions looked clearly better at it (an informal judgment, not a sweep). So that the encoder still produces embeddings when inputs are missing, training zeroes time and location together for $20\%$ of 32-chip cubes. For the remaining chips it zeroes all bands with probability $0.10$ and half of the bands with probability $0.20$. Users without time or location pass zeros at inference. To give all sensors one shared target while keeping spectral and spatial detail in the patch tokens, it combines unnormalized masked-patch $L_1$ reconstruction ($\div 10$ on MODIS) with a teacher cosine representation loss on \texttt{CLS} in a $90{:}10$ ratio:
{\small
\begin{equation}
\mathcal{L} = 0.9\,\mathcal{L}_{\mathrm{MAE}\text{-}L_1}(\hat{\bm{X}}_{\mathrm{m}}, \bm{X}_{\mathrm{m}}) + 0.1\,\mathcal{L}_{\mathrm{rep}}(\bm{z}_{\mathrm{CLS}}, \bm{t}_{\mathrm{teach}}),
\label{eq:loss}
\end{equation}}%
The frozen teacher receives pseudo-RGB views (VV, VH, and their mean for Sentinel-1), resized to $518\times 518$ in Stage~2. Training used AdamW~\cite{loshchilov2019adamw} in \texttt{bf16}, with cosine annealing and warm restarts~\cite{loshchilov2017sgdr} every $5{,}000$ steps that cycle the learning rate between $5\times 10^{-6}$ and $5\times 10^{-4}$. \textbf{Stage~1} (Aug--Oct 2024, $78$ epochs) distilled \texttt{CLS} toward pooled SAM ViT-B/16~\cite{kirillov2023sam} features through \textbf{7 nested Matryoshka heads} (\texttt{dolls=[16, 32, 64, 128, 256, 768, 1024]})~\cite{kusupati2022matryoshka}, so that prefixes of \texttt{CLS} could be used alone for compact storage and search. We reconstructed the Stage~1 settings from the public code history. \textbf{Stage~2} (Oct--Nov 2024) switched the teacher to DINOv2-L/14 with registers~\cite{darcet2024registers}. It replaced \texttt{MRLLoss} and its Matryoshka heads with a training-time dense \texttt{nn.Linear(1024, 1024)} projection head $\bm{P}$, discarded at inference; we found that $\bm{P}$ stabilized full-$1{,}024\text{-D}$ probes. We selected the Stage~2 checkpoint at epoch~7 by validation loss. Stage~2 kept Stage~1's variance spike but lost its prefix ordering (\S\ref{sec:lesson_mrl}, Lesson~3).

\textbf{Compute, carbon, and Apache-2.0 relicensing.} Pretraining and catalog exports ran on AWS EC2 in \texttt{us-east-2} (Ohio), on 8-GPU L4, A10G, and L40S instances (\texttt{g6/g5/g6e.48xlarge}). Training ran on a Slurm cluster of these EC2 instances, with PyTorch Lightning DDP across the 8-GPU nodes, all reading the shared FSx for Lustre store. The model fit on one GPU at 32 chips per step, so plain data parallelism was enough. From August to November 2024 the account logged $42{,}031$ EC2 node-hours. The account's reported emissions across all services were $33.56\,\text{tCO}_2\text{e}$ location-based and $0.047\,\text{tCO}_2\text{e}$ market-based, from the AWS Customer Carbon Footprint data export (carbon model v1.0.0). EC2 in \texttt{us-east-2} accounts for $30.59\,\text{tCO}_2\text{e}$ of that ($0.040\,\text{tCO}_2\text{e}$ market-based). Market-based figures follow the GHG Protocol market-based method, which accounts for renewable energy Amazon purchases. These totals cover v1.5 pretraining, tests, and embedding-catalog inference together.

In May~2025 the project relicensed all Clay weights from OpenRAIL-M~\cite{contractor2022rail} to \textbf{Apache-2.0} (documentation CC-BY-4.0). The project's stated reasons (Clay-foundation/model issue~349) were less adoption friction than a non-standard license with use-based restrictions and downstream obligations, consistency with the Apache-2.0 code base, and an explicit patent grant, which CC-BY-4.0 does not provide. Custom licenses also leave open questions that standard licenses avoid: where a license bars military or extractive uses, for example, it is unclear whether disaster response by a defense agency or a regulator's monitoring of a mining concession falls under them.

\section{Evaluation}
\label{sec:comparative}

\textbf{Table~\ref{tab:master_bench}} compares the frozen Clay~v1.5 checkpoint with other EO and general-purpose models on 43 benchmarks. Panel~A comes from the SatML leaderboard~\cite{taylorgeospatial2026satml}, an independent harness that runs about 150 models under one protocol: linear probes on 15 classification datasets, and a UPerNet head on three segmentation datasets. Clay's segmentation scores are the harness maintainers' own runs. Panel~B combines our own runs and published peer results on benchmarks that peer models report and the leaderboard does not cover: the AlphaEarth/TESSERA suite (all 15 tasks), detection and segmentation tasks from the OlmoEarth suite~\cite{herzog2026olmoearth,olmoearth2026v13}, and sub-meter, SAR, and unseen-sensor tasks from GEO-Bench~\cite{lacoste2023geobench}, Copernicus-Bench, and TorchGeo~\cite{stewart2022torchgeo}. Each row names the best other EO model and the best general-purpose model.

\subsection{Leaderboard and Single-Date Results}
\label{sec:std_bench}

On our own single-date tasks (Table~\ref{tab:master_bench}, Panel B), Clay and the baselines we ran follow one protocol: official splits, band standardization fitted on the training split after chip-mean nodata fill, and a validation-selected $\ell_2$ logistic/Ridge probe, or a zero-decoder $1\times 1$ linear segmentation probe, over $\{\texttt{CLS},\texttt{mean}\}$. Peer cells from~\cite{olmoearth2026v13} are third-party. Classification chips under $224\,\text{px}$ are resized to $224\times 224$~\cite{corley2024resize}. Together with the leaderboard, the results fall into three regimes.

(1) \textbf{Dense tokens may help very-high-resolution segmentation (Panel B).} With a frozen $1\times 1$ linear probe and no multi-scale decoder, Clay leads \texttt{m-pv4ger-seg} and \texttt{m-chesapeake} by $4\text{--}7$ mIoU over the best peer under the same probe. Its $8\times 8$ stride gives $4\times$ as many spatial tokens as a $16\times 16$ ViT ($723$ vs. $165\,\text{GFLOPs}$~\cite{sharma2026vhrbench}). With multi-scale decoders~\cite{xiong2024dofa,waldmann2025panopticon}, \texttt{DOFA-L} / \texttt{Panopticon} score higher on \texttt{m-pv4ger-seg} ($95.0/94.1$) but not on \texttt{m-chesapeake} ($66.3/60.8$), and reach $56.9/45.1$ on \texttt{m-cashewplant} (against $13.5/14.9$ with the linear probe).

(2) \textbf{Wavelength conditioning transfers across multispectral, SAR, and an unseen 21-band sensor (Panels A and B).} On the leaderboard's 11 Sentinel-2, Landsat, and Sentinel-1 classification datasets, Clay reaches $93\text{--}100\%$ of the best model. On four of them (\texttt{m-brick-kiln}, \texttt{forestnet}, \texttt{m-forestnet}, \texttt{treesatai}), its best entry uses RGB bands only. On 21-band Sentinel-3 OLCI (\texttt{Biomass-S3}, \texttt{LC100Cls-S3}), which it never saw in pretraining, it reaches $91\text{--}94\%$ of the best EO model. On the three segmentation datasets (Panel A), Clay with a frozen encoder reaches $94\text{--}100\%$ of the leaderboard cells, most of which are fine-tuned best runs.

(3) \textbf{RGB scene accuracy trails general backbones (Panel A).} On the four aerial RGB scene datasets (\texttt{aid}, \texttt{m-pv4ger}, \texttt{resisc45}, \texttt{ucmerced}), Clay reaches $92\text{--}98\%$ of the best model, behind general-purpose RGB backbones. A likely reason is that $0.9$ of the Stage-2 loss weight goes to reconstructing raw multi-sensor pixels. We also ran two models that are not on the leaderboard through the same harness on RGB input, and both show the same split. EmbeddingGemma~2~\cite{schechtervera2026embeddinggemma2}, a general multimodal embedder, scores $0.5\text{--}6$ points above Clay on these four datasets and $2\text{--}15$ points below it on the other 11. It ties the best general model on \texttt{aid}. FarSLIP~\cite{li2025farslip}, a ViT-B remote-sensing CLIP model, scores above Clay on these four and on \texttt{m-eurosat} and below it on the other 10. It is the best other EO model on \texttt{aid} and \texttt{resisc45}.

\subsection{Single-Date and Sensor-Agnostic}
\label{sec:aef_eval}

Table~\ref{tab:master_bench} highlights two trends in the ranking across the broader 2025--2026 evaluation suites: whether a model sees several dates, and whether it accepts any sensor. These suites are OlmoEarth-24, AlphaEarth/\allowbreak TESSERA-15, GEO-Bench-2, TerraTorch, and VHR-Bench; Table~\ref{tab:master_bench} covers part of them. In the \textbf{OlmoEarth~v1.3} and community suites, multi-temporal sequence encoders (OlmoEarth~v1.3, Galileo, TerraMind, Prithvi-EO-2.0) lead when the target depends on multi-month crop phenology (\texttt{PASTIS}, \texttt{Nandi}, \texttt{Breizh\-Crops}~\cite{olmoearth2026v13}). Dedicated optical models lead on pure-RGB texture (DINOv3 on VHR-Bench~\cite{sharma2026vhrbench}). Fixed-schema models cannot ingest sub-meter orthoimagery, $21$-band Sentinel-3, or custom band subsets without dropping channels or mapping unsupported bands into surrogate Sentinel-2 slots (\texttt{Olmo\-Earth}$^\natural$). Clay~v1.5 takes each task's own sensor. It is one of four models, with AnySat, Panopticon, and OlmoEarth, that run all $24/24$ frozen tasks in~\cite{olmoearth2026v13}. External 5-shot flood mapping across S1, S2, and PlanetScope agrees~\cite{kaushik2026flood}.

The AlphaEarth/TESSERA rows of Panel B compare single-date ViTs with precomputed $10\,\text{m}$ annual composites (TESSERA, AlphaEarth) on all 15 AlphaEarth/TESSERA tasks ($128\,\text{px}$, $5$ seeds, $100\,\text{km}$ block bootstrap). This comparison shows two further effects. First, pooling the central $2\times 2$ patch tokens ($160\times 160\,\text{m}$ around each labeled parcel) beats full-chip mean pooling at 10-shot for \texttt{Clay} on $11$ of $15$ tasks and for \texttt{OlmoEarth~v1.3} on $14$ of $15$. It is mixed for \texttt{Prithvi-EO-2.0} ($8$ of $15$; \S\ref{sec:lesson_tokens}, Lesson~2). This readout is post hoc; \texttt{CLS} was pre-specified. With it, single-date Clay has higher 10-shot point estimates than \texttt{DOFA-L}, \texttt{Prithvi-EO-2.0}, and \texttt{DINOv2-L} on $13\text{--}14$ of $15$ tasks, and the intervals exclude zero on $9\text{--}11$. Single-date \texttt{OlmoEarth~v1.3} is above Clay on $14$ of $15$ tasks ($9$ by interval).

Second, 12-month representations hold a clear advantage on seasonal crop and land-cover mapping: \texttt{OlmoEarth~v1.3}$^\diamond$, \texttt{TESSERA}, and \texttt{AlphaEarth}, plus native $64\text{-D}$ AlphaEarth on urban indicators~\cite{gong2026urban}. This is consistent with a single overpass not separating cover types whose phenology diverges across a growing season. Averaging $4$ quarterly S2 dates, with \texttt{CLS} fixed, closes a median $10\%$ of the gap to AlphaEarth at \texttt{Max} on the five tasks where the gap is at least $0.05$ ($20\%$ with concatenation). Annual composites do not apply (\texttt{n/a}) to disaster response (\texttt{Sen1Floods11}, \texttt{Vessel Detect.}). On monthly water flux (\texttt{OpenET}), the annual AlphaEarth and TESSERA samples we joined give negative $R^2$. But the published month-aligned AlphaEarth value ($0.58$) is above Clay ($0.45$).

Clay~v1.5 today appears stronger when a workflow needs one permissive Apache-2.0 checkpoint, or inputs that fixed-schema or annual systems cannot take: unseen instruments such as $21$-band Sentinel-3 OLCI, sub-meter aerial imagery with non-RGB bands, or incomplete band subsets. For multi-month phenology tasks, a multi-temporal model or annual composite is the better choice, and for RGB scene classification, a general RGB backbone such as DINOv3.

\section{Four Lessons from Two Years of Use}
\label{sec:lessons}

We designed Clay~v0, released in early 2024, as a masked autoencoder for end-to-end fine-tuning, not as a standalone embedding model. In practice, embeddings have become one of the primary use cases for the Clay model. Chip embeddings are small: a $1{,}024$-D float16 vector takes $2\,\text{KB}$, against $1.3\,\text{MB}$ for the 10-band int16 $256\times 256$ image chip it summarizes (${\sim}640\times$ smaller), so screening large areas or long time series can run on stored vectors instead of imagery ~\cite{gomes2024nec,vinge2025neucobench}. We do not compare chip-level and pixel-level accuracy per unit cost here. Public GeoParquet catalogs on Source Cooperative (\href{https://source.coop/clay}{\texttt{https://\allowbreak source.\allowbreak coop/\allowbreak clay}})~\cite{clay2026sourcecoop} now serve over $15.2\,\text{B}$ global Sentinel-2 Clay embeddings across $2.8\,\text{M}$ scenes (2017--2026, \texttt{clay/\allowbreak lgnd-\allowbreak embeddings}) and CONUS-wide 2010--2021 NAIP~v1.5 embeddings (\texttt{clay/\allowbreak clay-\allowbreak v1-\allowbreak 5-\allowbreak naip-\allowbreak 2}). The NAIP catalog was chipped with \texttt{stacchip}.

These lessons draw in part on work by the authors applying Earth embeddings on over $20$ projects in which we partnered with nonprofits, companies and a development bank to turn open satellite data and AI into working answers to their questions, taking each from a first conversation through discovery and prototype toward a build they could run themselves~\cite{grimshaw2026soy,chen2026aquaculture}. Partners include Amazon Sustainability, Aquarry, the Asian Development Bank, Carbon Mapper, CTrees, the Jane Goodall Institute, The Nature Conservancy, Seafood Watch, Trase, and Xylo Earth.

\paragraph{1. Frozen backbones with spatial decoders come within $0.5$ IoU of full fine-tuning on flood segmentation; a from-scratch U-Net is within $0.7$ of both.}
\emph{Takeaway:} keep the backbone frozen and train a small spatial decoder when boundaries matter; a per-token linear head is too coarse at $10\,\text{m}$. \label{sec:lesson_finetune}
On 10-band Sentinel-2 flood segmentation (\texttt{Sen1Floods11}~\cite{bonafilia2020sen1floods11}), a $1\times 1$ linear head on frozen $80\,\text{m}$ tokens blurs $10\,\text{m}$ shorelines ($74.8$ water IoU vs.\ $83.1$ for a $31\text{M}$-parameter from-scratch U-Net). A lightweight UPerNet decoder~\cite{xiao2018upernet} ($8.9\text{M}$ params) on the \textbf{frozen} Clay encoder raises water IoU to $83.3$ with $1/36$ of the trainable parameters. Full fine-tuning ($320.3\text{M}$ parameters) reaches $83.8$ (difference $+0.5$, paired $95\%$ interval $[-0.1, +1.5]$). Consistent with PANGAEA~\cite{marsocci2026pangaea}, foundation models barely edge out a supervised U-Net at $100\%$ labels. With few labels, others report clear gains: on 5-shot flood mapping~\cite{kaushik2026flood}, and on $11$-class tropical land-cover segmentation, where frozen Clay~v1 with a decoder gains $+7.5\text{--}11.4$ mIoU over U-Net++ and Prithvi~\cite{wiratama2025indonesia}.

\paragraph{2. \texttt{CLS} summarizes the scene, patch tokens keep the layout, and sensors do not share a common semantic space.}
\emph{Takeaway:} use \texttt{CLS} for scene-level search and classification, patch tokens when position within the chip matters, and do not expect a probe trained on one sensor to transfer to another. \label{sec:lesson_tokens}
Pooling trade-offs~\cite{corley2026pooling} are sharper in Clay~v1.5 because Stage~2 distilled \texttt{CLS} toward an RGB teacher while supervising the patch tokens $\bm{Z}_{\mathrm{patch}}$ through multi-band pixel reconstruction (Figure~\ref{fig:arch}). The patch tokens retain local radiometry and sub-chip layout. They recover within-chip spectral indices ($R^2 \approx 0.90$), solve $2\times 2$ quadrant localization that \texttt{CLS} mostly misses ($99\%$ vs.\ $42\%$; chance $25\%$), and beat broadcast \texttt{CLS} by $+6\text{--}16$ macro-$F_1$ points on ESA WorldCover across Sentinel-1/2, Landsat, and NAIP. Central $160\,\text{m}$ patch tokens likewise beat full-chip mean pooling for Clay and OlmoEarth (Table~\ref{tab:master_bench}, Panel~B).

\texttt{CLS} summarizes the scene. In USPVDB solar search~\cite{fujita2023uspvdb}, mean pooling dilutes compact arrays into background scrubland ($\text{AP} = 0.02$). \texttt{CLS} on $256\times 256$ chips preserves detection ($\text{AP} = 0.27$), as long as chips are not tightly cropped~\cite{corley2024resize}.

We identified three further challenges. (i)~\textbf{Shared weights do not align sensors into a common space.} In our own test across $6{,}592$ co-located CONUS sites on $320\,\text{m}$ footprints ($480\,\text{m}$ for Landsat), zero-shot transfer of a Sentinel-2 \texttt{CLS} probe to NAIP, Landsat, or Sentinel-1 lowers macro-$F_1$ by $41\text{--}57$ points. A $1{,}000$-pair Procrustes rotation recovers $0.19\text{--}0.40$ macro-$F_1$, against $0.56\text{--}0.83$ native. Chip size contributes: Sentinel-2 to NAIP drops $22$ points at $2.56\,\text{km}$ and $59$ at $320\,\text{m}$ (see also~\cite{tamazyan2025geocrossbench,marsocci2025xstars}). (ii)~\textbf{Clay treats time as frozen single-date snapshots}, with only a $[\sin,\cos]$ timestamp prior, so phenology and change semantics must be computed post hoc as differences or derivatives across embedding time series. (iii)~\textbf{Clay lacks native vision--language alignment}, so open-vocabulary retrieval and descriptive tasks require fitting a post-hoc text bridge or supervised probe.

\paragraph{3. Do not slice the first dimensions.}
\emph{Takeaway:} keep all $1{,}024$ dimensions; on our NLCD probe, the first 64 do no better than 64 random ones. \label{sec:lesson_mrl}
In Stage~1, $7$ nested Matryoshka heads (\texttt{dolls=[16..1024]}) forced the first $16$ coordinates of \texttt{CLS} to carry $8.0\times$ higher variance than trailing dimensions. Stage~2 replaced those heads with a training-time \texttt{nn.Linear(1024, 1024)} projection $\bm{P}$ (\S\ref{sec:lineage}; Figure~\ref{fig:arch}). This left Stage~1's $8.0\times$ variance spike in \texttt{dims[0:16]} but rotated the discriminative features across all $1{,}024$ dimensions. On spatially blocked NLCD regression over $223$ Texas scenes~\cite{dewitz2023nlcd}, naive prefix slicing (\texttt{dims[0:64]}) drops $R^2$ from $0.83$ (full $1{,}024\text{-D}$) to $0.62$, the same as $64$ random coordinates. \texttt{CLS} also linearly encodes its own per-sample pretraining reconstruction loss, so a Ridge probe recovers it without running the decoder (\textbf{ELLE}~\cite{nuno2026elle}).

\paragraph{4. Regularized parametric probes (LSVM, logistic/Ridge, 2-layer MLP) beat centroids, $k$-NN, trees, and 2D UMAP.}
\emph{Takeaway:} start with a linear SVM or logistic regression on the embeddings before trying nearest neighbors, trees, or 2D projections. \label{sec:lesson_probes}
On spatially blocked EuroSAT~\cite{helber2019eurosat} ($50\text{--}200$ labels/class), regularized parametric probes (squared-hinge LSVM~\cite{cortes1995svm,fan2008liblinear}, $\ell_2$ logistic regression, and $2$-layer MLPs) all reach $91\text{--}94\%$ macro-$F_1$. They beat $\ell_2$-normalized RandomForest and HistGBDT by $3\text{--}5$ points, cosine $k$-NN by $5$ points, class centroids by $8$ points, and train-only 2D UMAP~\cite{mcinnes2018umap} $+ k$-NN by $8\text{--}13$ points. On continuous NLCD canopy regression, 2D UMAP also drops $R^2$ from $0.85\text{--}0.87$ to $0.62$. In imbalanced $1{:}10$ concept search under $30\%$ label noise, $2$-layer MLPs and LSVMs ($\sim 0.90\text{ AP}$) also beat nearest-neighbor Max-Sim~\cite{johnson2021faiss} ($0.37\text{ AP}$) by learning negative-class boundaries.

\section{Conclusion}
\label{sec:conclusion}

Across $43$ benchmarks, two years after Clay~v1.5's release, specialized 2025--2026 models lead: optical models (DINOv3) on pure-RGB benchmarks, and multi-temporal Sentinel models on phenology benchmarks. Clay~v1.5 leads two $0.1\text{--}1\,\text{m}$ segmentation tasks under a frozen linear probe. It reaches $91\text{--}94\%$ of the best EO model on an unseen $21$-band instrument and a $98\%$ median of the best model on an independent leaderboard's $15$ classification datasets. It is a permissive Apache-2.0 encoder for sub-meter, multi-band, or unseen-instrument inputs without retraining. We draw three design recommendations for next-generation sensor-agnostic EO models. (1)~Pair dynamic wavelength stems with explicit cross-sensor and vision--language alignment for open-vocabulary retrieval and descriptive tasks, and model multi-date sequences natively rather than reducing time semantics to post-hoc derivatives of single-snapshot embeddings. (2)~Decouple attention stride from dense readout resolution via sub-patch poolers~\cite{olmoearth2026v13}. (3)~Retain Matryoshka supervision through the final distillation stage.

{\small
\paragraph{Data, Code, and Acknowledgments.}
\label{sec:acknowledgments}
Apache-2.0 code (\path{Clay-foundation/model}, \path{Clay-foundation/stacchip}), Apache-2.0 weights (\path{huggingface.co/made-with-clay/Clay}), CC-BY-4.0 embedding catalogs (\path{source.coop/clay}), and companion evaluation scripts (\path{Clay-foundation/clay-v1.5-paper}) are publicly available. \textbf{Funding and competing interests.} Clay is a program of Renaissance Philanthropy. Clay v0 through Clay v1.5 were developed with support from Schmidt Futures, the Grantham Foundation for the Protection of the Environment, and Amazon Web Services (AWS) while Clay was a fiscally sponsored project of Radiant Earth (501(c)(3)). During that period, Ode contributed to model development, strategy, operations, and the development of applications and demos. Development Seed contributed to building and training the Clay models. Legend AI funded its authors' work on this report, including the evaluation. AWS, the Schmidt Family Foundation, the Monterey Bay Aquarium, and Amazon Sustainability also funded applied projects. Legend AI builds commercial products on Earth embeddings, including those produced from Clay models. M.G., D.H., A.C., and L.C. are affiliated with Ode, which does applied work with Earth embeddings. We thank \textbf{Wei Ji Leong}, \textbf{Lilly Thomas}, \textbf{Leo Thomas}, \textbf{Ryan Avery}, \textbf{Myscon Truong}, \textbf{Mansour Raad}, \textbf{Jeff Albrecht}, and \textbf{Nat Manning}. We thank \textbf{Romeo Kienzler} for his contributions to packaging Clay as a \texttt{pip}-installable Python library. We also thank open-source contributors Aimee Barciauskas, Dan Bonomo, Kevin Booth, Isaac Campbell, Michele Claus, Isaac Corley, Chuck Daniels, Beno\^it de Chateauvieux, Tyler Erickson, Aninda Goswamy, Riddhi Goswami, Ash Hoover, Kelsey Josund, Freddie Kalaitzis, Maxime Lenormand, Javier Marin Tur, Bill Morris, Beatriz Peres, Ra\'ul Ramos, Ferdinand Schenck, Peter Sherman, Jed Sundwall, Jonas Viehweger, Lauren Yee, and Brayden Zhang, and the TorchGeo and TerraTorch maintainers (as of November 2026). We thank Massimiliano Brighindi, Nils Lehmann, and Filip Wolf for corrections and an independent evaluation after the first version of this report.

\section*{Responsible AI Use Disclosure}
\label{sec:ai_disclosure}
In November and December 2026 the authors used Gemini Carbon, Claude Opus, Claude Fable, and ChatGPT Astra for research, drafting, and adversarial review. The authors take full responsibility for this work.
}

\FloatBarrier
\bibliographystyle{plainnat}
{\footnotesize\raggedright\setlength{\bibsep}{1.5pt plus 1.5pt minus 0.5pt}\bibliography{references}}

\end{document}